\documentclass{article}
\usepackage{ijcai24}

\usepackage{times}
\usepackage{soul}
\usepackage{url}
\usepackage[hidelinks]{hyperref}
\usepackage[utf8]{inputenc}
\usepackage{multirow} 
\usepackage[small]{caption}
\usepackage{algorithmic}

\usepackage[ruled,vlined,lined,ruled,linesnumbered]{algorithm2e}
\usepackage{graphicx}
\usepackage{amsmath}
\usepackage{amsthm}

\usepackage{booktabs}
\usepackage[table,xcdraw]{xcolor}
\usepackage{color}

\SetKwInput{KwInput}{Input}                
\SetKwInput{KwOutput}{Output}              

\newlength\mylen 

\title{Heterogeneous Federated Learning with Convolutional and Spiking Neural Networks}

\author{
Yingchao Yu$^{1}$
\and
Yuping Yan$^{2,3}$\And
Jisong Cai$^{4}$\And
Yaochu Jin$^{2,*}$\\
\affiliations
$^1$College of Information Science and Technology, Donghua University\\
$^2$School of Engineering, Westlake University\\
$^3$Faculty of Informatics, Department of Computeralgebra, Eötvös Loránd University\\
$^4$School of Cyber Science and Engineering, Wuhan University\\
$^*$ Corresponding author\\
\emails
yingchaoyuu@outlook.com,
\{yanyuping, jinyaochu\}@westlake.edu.cn,
\ 2021302181110@whu.edu.cn
}

\begin{document}

\maketitle

\begin{abstract}
Federated learning (FL) has emerged as a promising paradigm for training models on decentralized data while safeguarding data privacy. Most existing FL systems, however, assume that all machine learning models are of the same type, although it becomes more likely that different edge devices adopt different types of AI models, including both conventional analogue artificial neural networks (ANNs) and biologically more plausible spiking neural networks (SNNs). This diversity empowers the efficient handling of specific tasks and requirements, showcasing the adaptability and versatility of edge computing platforms. One main challenge of such heterogeneous FL system lies in effectively aggregating models from the local devices in a privacy-preserving manner. To address the above issue, this work benchmarks FL systems containing both convoluntional neural networks (CNNs) and SNNs by comparing various aggregation approaches, including federated CNNs, federated SNNs, federated CNNs for SNNs, federated SNNs for CNNs, and federated CNNs with SNN fusion. Experimental results demonstrate that the CNN-SNN fusion framework exhibits the best performance among the above settings on the MNIST dataset. Additionally, intriguing phenomena of competitive suppression are noted during the convergence process of multi-model FL.
\end{abstract}

\section{Introduction}

Embedded artificial intelligence and heterogeneous edge devices will be in increasing demand in various industrial and IoT systems. Alongside devices incorporating traditional convolutional neural networks (CNNs) \cite{rashid2022ahar,merenda2020edge}, those utilizing spiking neural networks (SNNs) \cite{koo2020sbsnn,yang2022lead} will also emerge as strong contenders due to their advantage of low power consumption. Effectively and securely leveraging data from these heterogeneous devices will be key to this scenario. 

Both CNNs and SNNs are artificial neural networks that can be used to solve various AI tasks, however, they have distinct architectures and operate in different ways. CNNs \cite{gu2018recent} are primarily used for tasks involving grid-structured data, such as images, videos, and signals. They are composed of multiple layers, including convolutional layers, pooling layers, and fully connected layers. CNNs are widely used in image recognition, object detection, image segmentation, and other computer vision tasks due to their ability to capture spatial dependencies in data efficiently.

SNNs \cite{ghosh2009spiking} are a type of neural network model inspired by biological neurons, particularly in their use of spikes or action potentials for information processing. Unlike CNNs, SNNs operate in discrete time steps, with neurons firing spikes in response to input stimuli. SNNs can process temporal information efficiently and are well-suited for tasks involving spatiotemporal data, such as time series prediction, speech recognition, and sensor data processing. Due to its event-driven and energy-efficient nature, SNNs are more suitable for mobile and edge devices. 

Given the distinct suitability of CNNs and SNNs for different scenarios, exploring the potential of combining their capabilities across various modalities presents a practical challenge for addressing their individual limitations. By integrating the diverse modal data processed by CNNs and SNNs into joint learning frameworks, we can leverage the complementary strengths of each architecture to enhance overall model performance and robustness. However, the establishment of a secure and privacy-preserving environment for CNN-SNN interaction is another critical challenge.

Federated learning (FL) \cite{ji2024emerging,mcmahan2017communication} has emerged as a promising solution to this problem, offering a decentralized approach to model training that respects data privacy and security. While existing federated CNN \cite{lu2020efficient} and federated SNN \cite{venkatesha2021federated,skatchkovsky2020federated,wang2023efficient} frameworks enable training within a single modality, the exploration of multi-modal learning techniques that leverage both SNN and CNN capabilities remains unclear. By bridging this gap and developing novel approaches for multi-modal learning, we can unlock the full potential of CNN-SNN fusion in  federated environment. 

In this paper, we present a novel approach to multi-modal federated learning, leveraging the fusion of CNNs and SNNs. The contributions of this study are as follows:
\begin{itemize}
    \item This paper pioneers the integration of CNN-SNN models within an FL framework. 
    \item To elucidate the efficacy of our proposed framework, we conduct a thorough comparative analysis against various federated learning approaches. Specifically, we compare our CNN-SNN fusion framework with federated CNNs, federated SNNs, federated CNNs for SNNs, and federated SNNs for CNNs. 
    \item Our CNN-SNN fusion outperforms the CNN for SNN and SNN for CNN frameworks and we observe interesting competitive suppression phenomena during the training process of CNN-SNN fusion. 
\end{itemize}

\section{Preliminaries}
In this section, we introduce the preliminaries relevant to our work, namely FL and SNNs.

\subsection{Federated Learning}
FL is a distributed machine learning paradigm in which each of the participants trains a model on local data and uploads the parameters of the updated model to the server. Then the server aggregates the local models to obtain a global model. Compared with traditional machine learning techniques, FL cannot only improve learning efficiency but also solve the problem of data silos and protect local data privacy \cite{custers2019eu}. 

In horizontal FL (HFL) \cite{yang2019federated,zhang2023federated}, the feature space is shared across all parties, but each party may have a distinct sample space in their datasets, representing one of the most prevalent frameworks in federated learning. Initially, the server initializes a model with random parameters $\theta_0$ and distributes it to all participating clients. Subsequently, a subset of $k$ out of $n$ clients receive the model and compute training gradients locally based on their respective datasets. These updated models are then transmitted back to the server, which aggregates the gradients from all participating clients to compute the global parameters:
\begin{equation}
\theta_r=\sum_{i=1}^{k}\theta_i/k
\end{equation}

Federated Averaging (FedAvg) \cite{mcmahan2017communication,wang2020federated} is one of the most widely adopted methods in FL. Its pseudocode is delineated in Algorithm \ref{alg:Fedavg}.

\begin{algorithm}
\caption{FedAvg}\label{alg:Fedavg}
\KwInput{The $\mathnormal{K}$ clients are indexed by $k$, $\mathnormal{B}$ is the local mini batch-size, $\mathnormal{E}$ is number of local epochs, and $\eta$ is the learning rate}
\KwOutput{Updated model}

Server Side:
weights initialization $w_o$\\
\textbf{do} \\
$\ \ \ \ \ $Select $S_t\leftarrow m\leftarrow(\mathnormal{C}*\mathnormal{K},1)$ clients randomly.\\
$\ \ \ \ \ \ $ \textbf{for} each client $k \in{S_t}$ \textbf{in parallel do}\\
\ \ \ \ \ \ \ \  $w_{t+1}^k \leftarrow ClientUpdate (k,w_t)$\\
$\ \ \ \ \ \ \ \ $ $w_{t+1}\leftarrow \sum_{k=1}^{K}\frac{n_k}{n}w_{t+1}^k$\\
$\ \ \ \ \ \ $\textbf{end for}\\
\textbf{end for}\\
\textbf{Return} $w_{t+1}$ \\
Local update$(k,w)$\\
$\mathcal{B} \leftarrow$ split $d_k$ into batches of size $\mathnormal{B}$\\
\textbf{for} each local epoch $e$ from 1 to $E$ \textbf{do} \\ 
$\ \ \ \ \ \ $\textbf{for} $b \in \mathcal{B}$ \textbf{do} \\
\ \ \ \ \ \ \ \  $w\leftarrow w-\eta \nabla (w,b)$ \\ 
$\ \ \ \ \ \ $ \textbf{end for}\\
\textbf{end for}\\
\textbf{Return} $w$ \\

\end{algorithm}

In this method, stochastic gradient descent (SGD) \cite{johnson2013accelerating} is utilized to minimize the global dropout and accuracy function $f_{FL}$. In Federated Stochastic Gradient Descent (FedSGD)\cite{mcmahan2017communication}, all clients are involved ($C=1$), and each client $k$ updates its local parameters as follows: 
\begin{equation}
    w_{t+1}^k \leftarrow w_t-\nabla g_k.
\end{equation}
Subsequently, the server aggregates these updated parameters into a global parameter $w_{t+1}$ using weighted averaging: 
\begin{equation}
    w_{t+1}\leftarrow \sum_{k=1}^{K}\frac{n_k}{n}w_{t+1}^k
\end{equation}
If clients conduct multiple updates within their local datasets, the local parameters are iteratively updated by: 
\begin{equation}
    w^k \leftarrow w^k -\eta \nabla F_k(w_k).
\end{equation}

\begin{figure}[htb]
    \centering
    \includegraphics[width=3in]{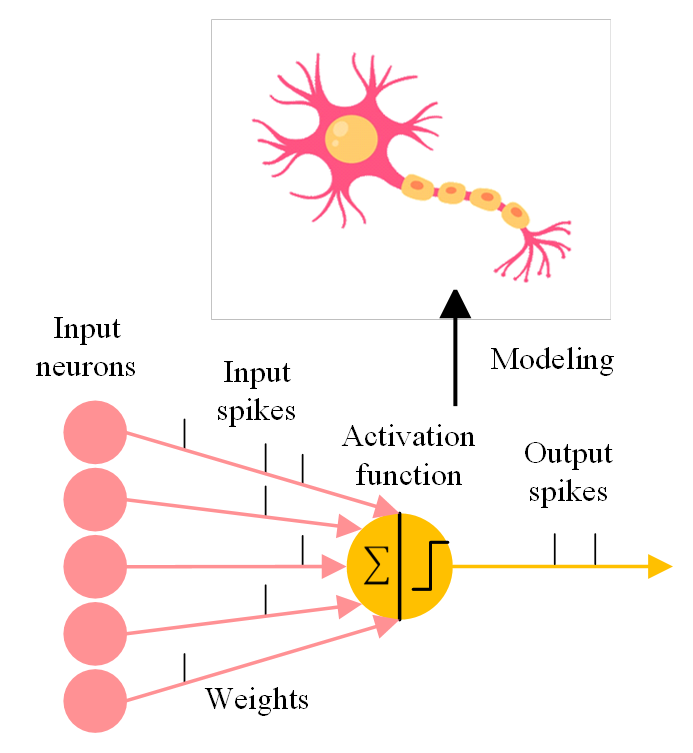}
    \caption{An illustration of behavior of a spiking neuron.}
    \label{snn}
\end{figure}

\begin{figure}[htb]
    \centering
    \includegraphics[width=3.5in]{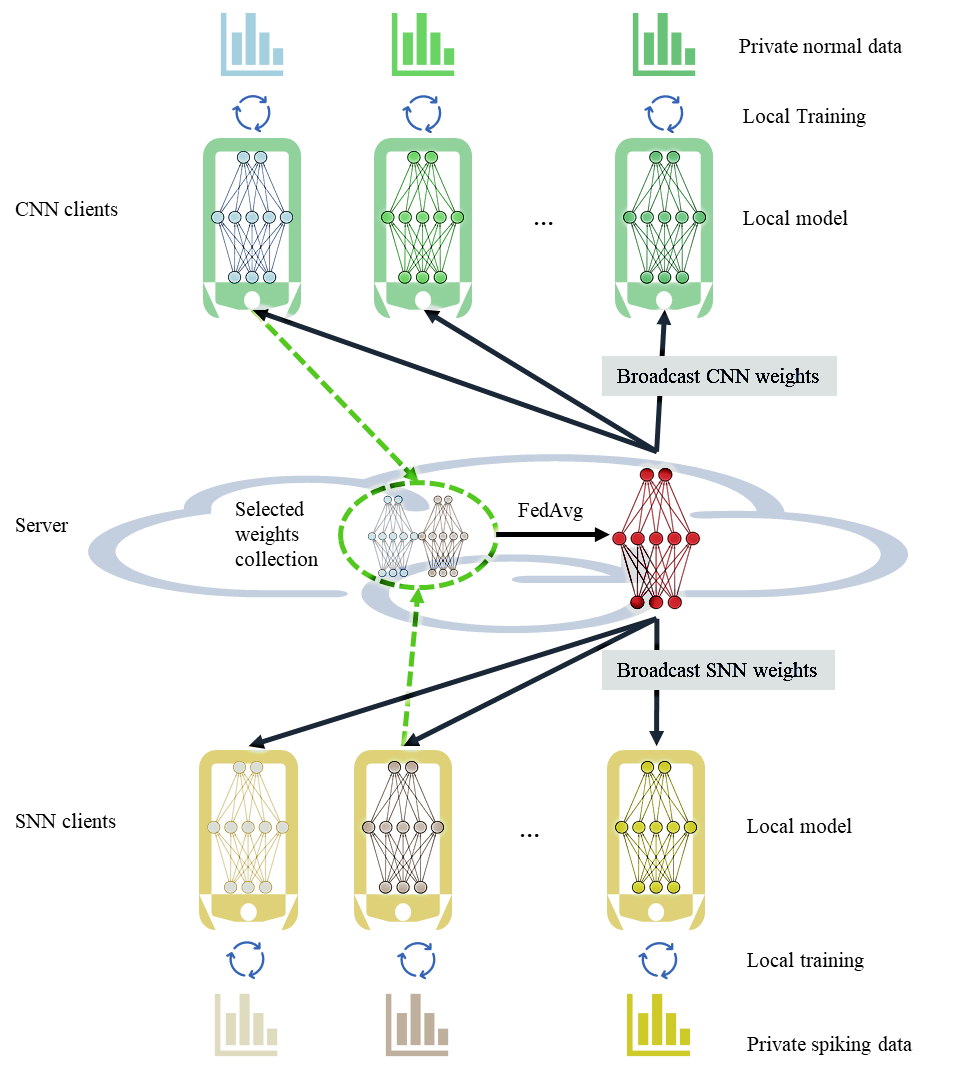}
    \caption{The federated CNN-SNN fusion framework.}
    \label{cnnsnn}
\end{figure}

\begin{figure}[htb]
    \centering
    \includegraphics[width=3.5in]{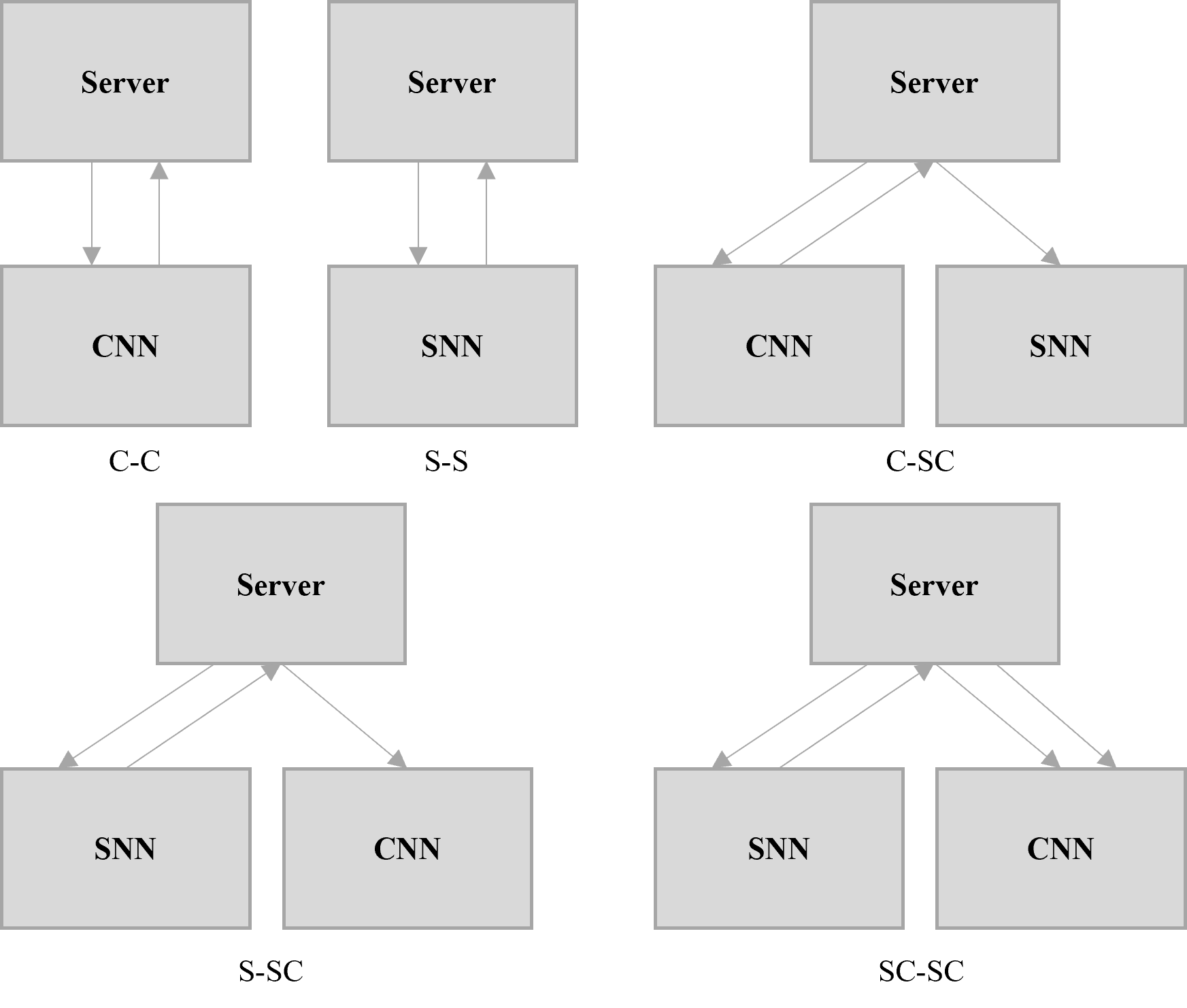}
    \caption{Five experimental frameworks.}
    \label{all}
\end{figure}

\begin{table*}[htbp]
\centering
\begin{tabular}{@{}|
>{\columncolor[HTML]{EEEEEE}}c 
>{\columncolor[HTML]{EEEEEE}}c 
>{\columncolor[HTML]{EEEEEE}}c |
>{\columncolor[HTML]{F0F0EC}}c 
>{\columncolor[HTML]{F0F0EC}}c 
>{\columncolor[HTML]{F0F0EC}}c |@{}}
\multicolumn{3}{|c|}{\cellcolor[HTML]{D5DCE4}\textbf{CNN}}                                                                                                                                                                      & \multicolumn{3}{c|}{\cellcolor[HTML]{D5DCE4}\textbf{SNN}}                                                                                                                                                                      \\
\multicolumn{1}{|c|}{\cellcolor[HTML]{BDD6EE}\textbf{Layer name}} & \multicolumn{1}{c|}{\cellcolor[HTML]{BDD6EE}\textbf{Param shape}}                                           & \cellcolor[HTML]{BDD6EE}\textbf{Output shape} & \multicolumn{1}{c|}{\cellcolor[HTML]{BDD6EE}\textbf{Layer name}} & \multicolumn{1}{c|}{\cellcolor[HTML]{BDD6EE}\textbf{Param shape}}                                           & \cellcolor[HTML]{BDD6EE}\textbf{Output shape} \\
\multicolumn{1}{|c|}{\cellcolor[HTML]{EEEEEE}Input}               & \multicolumn{1}{c|}{\cellcolor[HTML]{EEEEEE}-}                                                              & None×1×32×32                                  & \multicolumn{1}{c|}{\cellcolor[HTML]{F0F0EC}Input}               & \multicolumn{1}{c|}{\cellcolor[HTML]{F0F0EC}-}                                                              & None×T×1×32×32                                \\
\multicolumn{1}{|c|}{\cellcolor[HTML]{EEEEEE}Conv2d}              & \multicolumn{1}{c|}{\cellcolor[HTML]{EEEEEE}Weights: 32×1×3×3}                                              & None×32×32×32                                 & \multicolumn{1}{c|}{\cellcolor[HTML]{F0F0EC}Conv2d}              & \multicolumn{1}{c|}{\cellcolor[HTML]{F0F0EC}Weights: 32×1×3×3}                                              & None×T×32×32×32                               \\
\multicolumn{1}{|c|}{\cellcolor[HTML]{EEEEEE}BatchNorm2d}         & \multicolumn{1}{c|}{\cellcolor[HTML]{EEEEEE}\begin{tabular}[c]{@{}c@{}}Weights: 32\\ Bias: 32\end{tabular}} & None×32×32×32                                 & \multicolumn{1}{c|}{\cellcolor[HTML]{F0F0EC}BatchNorm2d}         & \multicolumn{1}{c|}{\cellcolor[HTML]{F0F0EC}\begin{tabular}[c]{@{}c@{}}Weights: 32\\ Bias: 32\end{tabular}} & None×T×32×32×32                               \\
\multicolumn{1}{|c|}{\cellcolor[HTML]{EEEEEE}Sigmoid}             & \multicolumn{1}{c|}{\cellcolor[HTML]{EEEEEE}-}                                                              & None×32×32×32                                 & \multicolumn{1}{c|}{\cellcolor[HTML]{F0F0EC}IFNode}              & \multicolumn{1}{c|}{\cellcolor[HTML]{F0F0EC}-}                                                              & None×T×32×32×32                               \\
\multicolumn{1}{|c|}{\cellcolor[HTML]{EEEEEE}MaxPool2d}           & \multicolumn{1}{c|}{\cellcolor[HTML]{EEEEEE}-}                                                              & None×32×16×16                                 & \multicolumn{1}{c|}{\cellcolor[HTML]{F0F0EC}MaxPool2d}           & \multicolumn{1}{c|}{\cellcolor[HTML]{F0F0EC}-}                                                              & None×T×32×16×16                               \\
\multicolumn{1}{|c|}{\cellcolor[HTML]{EEEEEE}Conv2d}              & \multicolumn{1}{c|}{\cellcolor[HTML]{EEEEEE}Weights: 32×1×3×3}                                              & None×32×16×16                                 & \multicolumn{1}{c|}{\cellcolor[HTML]{F0F0EC}Conv2d}              & \multicolumn{1}{c|}{\cellcolor[HTML]{F0F0EC}Weights: 32×1×3×3}                                              & None×T×32×16×16                               \\
\multicolumn{1}{|c|}{\cellcolor[HTML]{EEEEEE}BatchNorm2d}         & \multicolumn{1}{c|}{\cellcolor[HTML]{EEEEEE}\begin{tabular}[c]{@{}c@{}}Weights: 32\\ Bias: 32\end{tabular}} & None×32×16×16                                 & \multicolumn{1}{c|}{\cellcolor[HTML]{F0F0EC}BatchNorm2d}         & \multicolumn{1}{c|}{\cellcolor[HTML]{F0F0EC}\begin{tabular}[c]{@{}c@{}}Weights: 32\\ Bias: 32\end{tabular}} & None×T×32×16×16                               \\
\multicolumn{1}{|c|}{\cellcolor[HTML]{EEEEEE}Sigmoid}             & \multicolumn{1}{c|}{\cellcolor[HTML]{EEEEEE}-}                                                              & None×32×16×16                                 & \multicolumn{1}{c|}{\cellcolor[HTML]{F0F0EC}IFNode}              & \multicolumn{1}{c|}{\cellcolor[HTML]{F0F0EC}-}                                                              & None×T×32×16×16                               \\
\multicolumn{1}{|c|}{\cellcolor[HTML]{EEEEEE}MaxPool2d}           & \multicolumn{1}{c|}{\cellcolor[HTML]{EEEEEE}-}                                                              & None×32×8×8                                   & \multicolumn{1}{c|}{\cellcolor[HTML]{F0F0EC}MaxPool2d}           & \multicolumn{1}{c|}{\cellcolor[HTML]{F0F0EC}-}                                                              & None×T×32×8×8                                 \\
\multicolumn{1}{|c|}{\cellcolor[HTML]{EEEEEE}Linear}              & \multicolumn{1}{c|}{\cellcolor[HTML]{EEEEEE}Weights: 512×2048}                                              & None×512                                      & \multicolumn{1}{c|}{\cellcolor[HTML]{F0F0EC}Linear}              & \multicolumn{1}{c|}{\cellcolor[HTML]{F0F0EC}Weights: 512×2048}                                              & None×T×512                                    \\
\multicolumn{1}{|c|}{\cellcolor[HTML]{EEEEEE}Sigmoid}             & \multicolumn{1}{c|}{\cellcolor[HTML]{EEEEEE}-}                                                              & None×512                                      & \multicolumn{1}{c|}{\cellcolor[HTML]{F0F0EC}IFNode}              & \multicolumn{1}{c|}{\cellcolor[HTML]{F0F0EC}-}                                                              & None×T×512                                    \\
\multicolumn{1}{|c|}{\cellcolor[HTML]{EEEEEE}Linear}              & \multicolumn{1}{c|}{\cellcolor[HTML]{EEEEEE}Weights: 10×512}                                                & None×10                                       & \multicolumn{1}{c|}{\cellcolor[HTML]{F0F0EC}Linear}              & \multicolumn{1}{c|}{\cellcolor[HTML]{F0F0EC}Weights: 10×512}                                                & None×T×10                                     \\
\multicolumn{1}{|c|}{\cellcolor[HTML]{EEEEEE}Sigmoid}             & \multicolumn{1}{c|}{\cellcolor[HTML]{EEEEEE}-}                                                              & None×10                                       & \multicolumn{1}{c|}{\cellcolor[HTML]{F0F0EC}IFNode}              & \multicolumn{1}{c|}{\cellcolor[HTML]{F0F0EC}-}                                                              & None×T×10                                     \\
\multicolumn{3}{|c|}{\cellcolor[HTML]{BDD6EE}-}                                                                                                                                                                                 & \multicolumn{3}{c|}{\cellcolor[HTML]{BDD6EE}Surrogate function of IFNode: sigmoid}                                                                                                                                            
\end{tabular}
\caption{Model architectures of CNN and SNN.}
\label{t2} 
\end{table*}

\subsection{Spiking Neural Networks}

In an SNN, information is encoded in the timing of these spikes \cite{lobo2020spiking}. The Leaky Integrate-and-Fire (LIF) \cite{kornijcuk2016leaky,hunsberger2015spiking} neuron model is one of the simplest and most commonly used neuron models in SNNs. It captures the basic behavior of biological neurons by integrating incoming input signals over time and emitting an output spike when a certain threshold is reached. 

An LIF neuron in layer $l$ with index $i$ can formally be described in differential form:
\begin{equation}
    \tau_{\mathrm{mem}} \frac{\mathrm{d} U_i^{(l)}}{\mathrm{d} t}=-\left(U_i^{(l)}-U_{\mathrm{rest}}\right)+R I_i^{(l)},
\end{equation}
where $U_i(t)$ is the membrane potential, $U_{\text {rest}}$ is the resting potential, $\tau_{\text {mem}}$ is the membrane time constant, $R$ is the input resistance, and $I_i(t)$ is the input current.

Figure \ref{snn} describes the basic calculation process of the SNN. Considering a neuron $i$ receiving input from a set of neurons $N$, the incoming spikes from these input neurons are weighted by the parameters $w_{ij}$ for all $j$ belonging to $N$ and are accumulated to form the neuron. Upon reaching a predefined threshold $v$, the neuron generates an output spike. Following this spike, the membrane potential undergoes a reset process. 

In SNNs, the input current is typically generated by synaptic currents induced by the arrival of presynaptic spikes $S_j^{(l)}(t)$. When working with differential equations, a spike train $S_j^{(l)}(t)$ as a sum of Dirac delta functions $S_j^{(l)}(t)=\sum_{s \in C_j^{(l)}} \delta(t-s)$ is denoted, where $s$ iterates over the firing times $C_j^{(l)}$ of neuron $j$ in layer $l$.

Synaptic currents exhibit distinct temporal dynamics. A common simplification is to model their time evolution as an exponentially decaying current following each presynaptic spike. Additionally, we assume linearity in the summation of synaptic currents. The dynamics of these operations can be encapsulated by the following expression:

\begin{equation}
    \frac{\mathrm{d} I_i^{(l)}}{\mathrm{d} t}=-\underbrace{\frac{I_i^{(l)}(t)}{\tau_{\text {syn }}}}_{\text {exp. decay }}+\underbrace{\sum_j W_{i j}^{(l)} S_j^{(l-1)}(t)}_{\text {feed-forward }}+\underbrace{\sum_j V_{i j}^{(l)} S_j^{(l)}(t)}_{\text {recurrent }}.
\end{equation}

Here, the summation extends over all presynaptic neurons $j$, with $W_{i,j}^{(l)}$ representing the corresponding afferent weights from the layer below. Additionally, $V_{i,j}^{(l)}$ denotes explicit recurrent connections $i j$ within each layer. We can represent a single LIF neuron through two linear differential equations, where the initial conditions undergo instantaneous modification upon each spike occurrence. Leveraging this property, we can introduce a reset term via an additional factor that immediately decreases the membrane potential by $\left(\vartheta-U_{\text {rest }}\right)$ whenever the neuron emits a spike:
\begin{equation}
    \frac{\mathrm{d} U_i^{(l)}}{\mathrm{d} t}=-\frac{1}{\tau_{\text {mem }}}\left(\left(U_i^{(l)}-U_{\text {rest }}\right)+R I_i^{(l)}\right)+S_i^{(l)}(t)\left(U_{\text {rest }}-\vartheta\right).
\end{equation}

\begin{table*}[htbp]
\centering
\begin{tabular}{llllllllll}
\multicolumn{2}{l}{\multirow{2}{*}{N/P:10/2}}               & \multicolumn{1}{c}{C-C} & \multicolumn{1}{c}{S-S}     & \multicolumn{2}{c}{C-SC} & \multicolumn{2}{c}{S-SC} & \multicolumn{2}{c}{SC-SC} \\ \cline{3-10} 
\multicolumn{2}{l}{}                                        & CNN                     & SNN                         & CNN             & SNN    & CNN    & SNN             & CNN             & SNN     \\ \hline
\multicolumn{2}{l|}{IID}                                    & 98.930                  & \multicolumn{1}{l|}{99.160} & 95.680          & 89.310 & 47.780 & 95.870          & \textbf{97.470} & 95.690  \\
\multirow{3}{*}{Non-IID} & \multicolumn{1}{l|}{Alpha=1}     & 98.950                  & \multicolumn{1}{l|}{98.610} & \textbf{97.740} & 93.570 & 54.240 & 97.560          & 95.6500         & 94.2700 \\
                         & \multicolumn{1}{l|}{Alpha=0.5}   & 98.570                  & \multicolumn{1}{l|}{97.890} & 97.900          & 92.770 & 45.300 & 97.360          & \textbf{97.500} & 90.710  \\
                         & \multicolumn{1}{l|}{Alpha=0.125} & 62.790                  & \multicolumn{1}{l|}{74.100} & 57.620          & 51.210 & 38.960 & \textbf{73.110} & 58.920          & 61.710  \\ \hline
\end{tabular}
\caption{Accuracy comparisons of different modes.}
\label{t1} 
\end{table*}

\begin{figure*}[htb]
    \centering
    \includegraphics[width=5in]{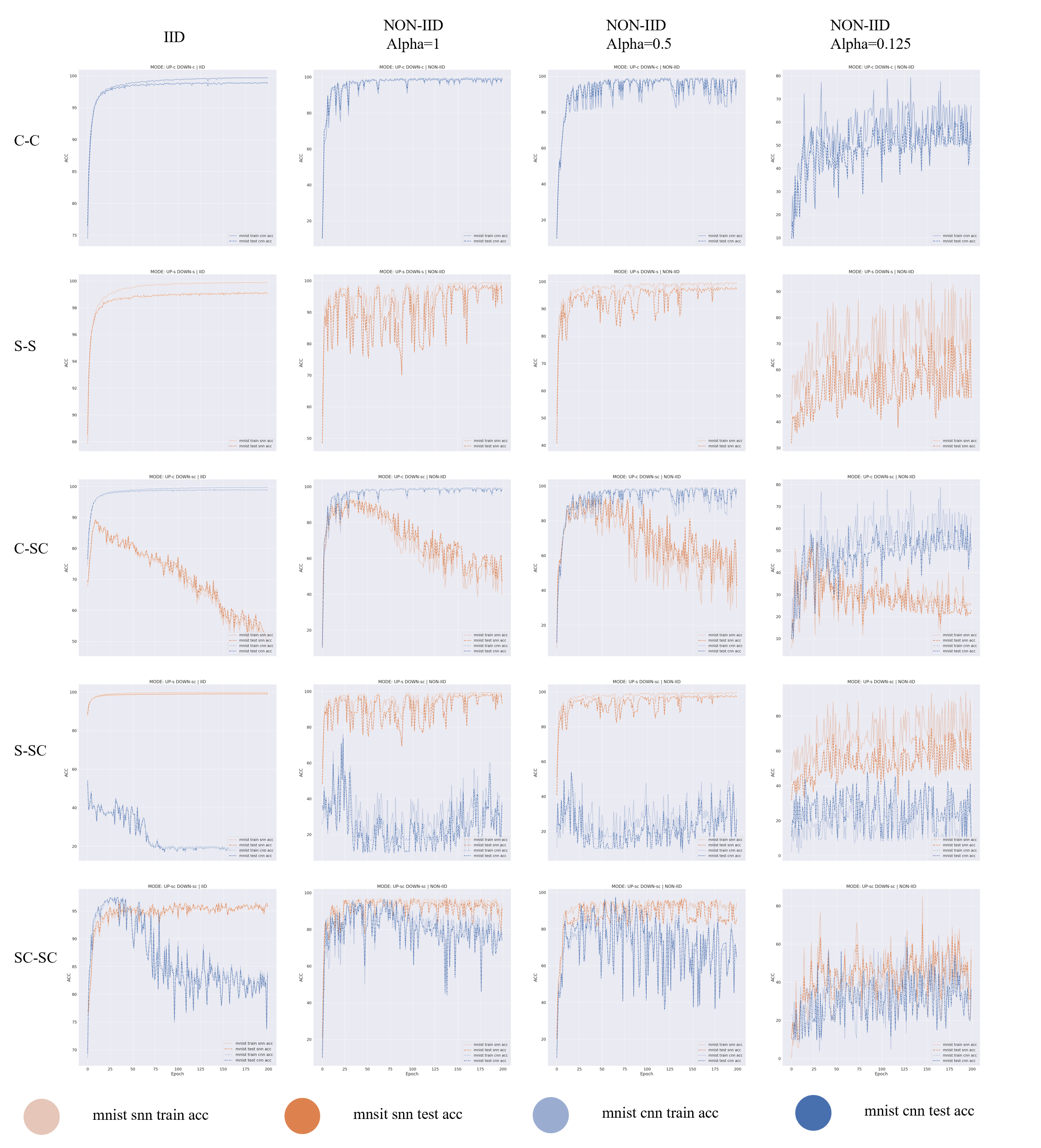}
    \caption{Accuracy of five frameworks under IID and Non-IID scenarios.}
    \label{exp}
\end{figure*}

\begin{figure*}[htb]
    \centering
    \includegraphics[width=6in]{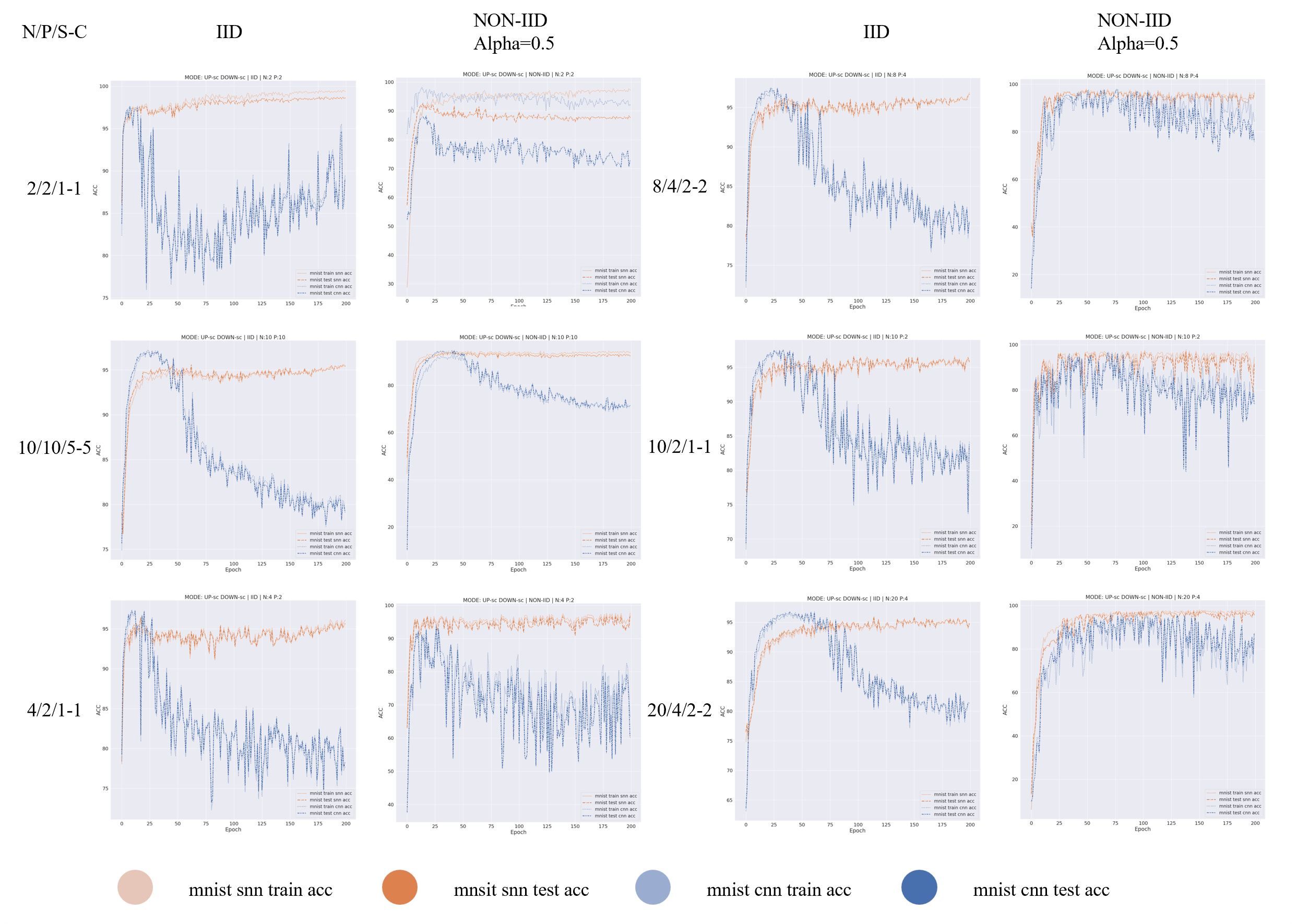}
    \caption{Accuracy of SC-SC framework for different numbers of clients.}
    \label{exp1}
\end{figure*}

\section{CNN-SNN Fusion Framework}

In the federated framework, a centralized server orchestrates multiple clients, each potentially equipped with either an SNN or a CNN model. The overarching goal is to train a global model without necessitating the transmission of raw data from the clients. It includes three phases, initialization, local training, and aggregation. 

\begin{itemize}
    \item \textbf{Initialization:}
    The federated learning process commences with the initialization phase, during which a global model is instantiated and distributed to all participating clients by the centralized server.

Each client receives the initial model parameters, serving as the foundational framework for subsequent training iterations.

    \item \textbf{Local training phase:} In the local training phase, individual clients train their respective local datasets to fine-tune the global model parameters. For SNNs, local training entails fine-tuning synaptic weights, thresholds, and other parameters critical for optimizing spike timings and firing rates. Employing the FedAvg algorithm, clients update model parameters and upload their parameters to the server.

    \item \textbf{Aggregation:} After receiving the parameters, the server conducts an aggregation with the average function and outputs the final result, which is the global model.

\end{itemize}

To fully demonstrate the effectiveness of the CNN-SNN fusion framework, we have devised five distinct modes, refer to Fig. \ref{all}, each tailored to specific functionalities and scenarios:
\begin{itemize}
    \item \textbf{Federated CNN Mode (C-C):} In this mode, the algorithm exclusively operates in the domain of CNNs. The training process encompasses all CNN clients, fostering a unified and optimized CNN-based approach across the network.

    \item \textbf{Federated SNN Mode (S-S):} By copntrast, the federated SNN mode focuses solely on SNNs. It orchestrates the training of all SNN clients.

    \item \textbf{Federated CNN for SNN Mode (C-SC):} This mode intertwines the modalities of both CNNs and SNNs. Initially, CNN models undergo training and are then uploaded to a centralized server. Then, all clients, regardless of whether they are using CNN or SNN, get these CNN weights from the server. This hybrid approach optimizes resources by leveraging CNN weights across the network.

    \item \textbf{Federated SNN for CNN Mode (S-SC):}  Similarly, the federated SNN for CNN mode combines CNN and SNN capabilities. Here, SNN models undergo training and are uploaded to the server. All clients, regardless of their inherent modality, then fetch these SNN weights from the server. This strategy optimizes network efficiency by capitalizing on SNN weights across diverse client devices.

    \item \textbf{CNN-SNN Fusion Mode (SC-SC):} The CNN-SNN fusion mode amalgamates the modalities of both CNNs and SNNs. Clients independently train their local models, contributing to a diverse and robust set of weights. The server orchestrates weight aggregation, culminating in a unified, comprehensive model. Subsequently, all clients retrieve these aggregated weights, embodying a fusion of CNN and SNN capabilities across the network.
\end{itemize}

\section{Experimental Results}
The experiments assess five distinct frameworks using the MNIST dataset \cite{deng2012mnist}. In each framework, an equal number of CNN and SNN clients undergo 200 global epochs, with each client performing 2 local epochs.

\subsection{Model Architectures and Data Processing}

Table \ref{t2} presents the architectures of the CNN and SNN models utilized in the experiments. The inputs to the SNN clients are temporal discrete pulse signals, generated from continuous signals with a temporal length set to 20 time steps and following a Poisson distribution \cite{heeger2000poisson}. The activation function in the SNN is based on the Integrate-and-Fire neuron model, corresponding to IFNode in Table \ref{t2}, which enables the SNN to process spiking signals. Furthermore, a sigmoid surrogate gradient function is employed to facilitate backpropagation training in the SNN.


The data for the CNN and SNN clients are randomly distributed, non-overlapping, and remain constant across epochs from the MNIST dataset. In non-IID scenarios, a Dirichlet distribution is employed to allocate sample proportions among different clients for each class, generating non-IID data \cite{zhu2021federated}, with the parameter alpha governing the degree of non-IIDness, where lower values indicate higher degrees of non-IIDness.


\subsection{Comparison of Five Frameworks}
In this series of experiments, we configured the number of clients to 10 and uploaded one SNN weight and one CNN weight each time. Additionally, the non-IID data set is further divided into three different levels of non-independence, represented by different alpha values. Lower alpha values indicate higher non-independence, which makes the task of generalizing the model more challenging.  

From the results in Figure \ref{exp}, we observe that the proposed CNN-SNN fusion framework (SC-SC) can converge under both IID and various degrees of non-IID data. Secondly, within the SC-SC framework, there exists a convergence difference between SNN and CNN due to the misalignment of different modal data. Their gradients compete during fusion, with the dominant side suppressing the performance of the weaker side during training. In this experiment, SNN clients are the dominant side. 

Additionally, compared to the C-SC and S-SC frameworks, the SC-SC framework exhibits smaller convergence differences between SNN and CNN. This is because the SC-SC framework merges the gradients of two modal models, allowing the server to simultaneously learn knowledge from both SNN and CNN clients, while C-SC and S-SC models can only learn from one type of client. Furthermore, as the degree of Non-IID data increases, the differences between CNN and SNN results gradually decrease in the C-SC, S-SC, and SC-SC frameworks. This indicates that non-IID data helps alleviate competition suppression issues in training different modalities.

Finally, compared to the C-C and S-S frameworks, the performance of the SC-SC framework on each modality is slightly lower than the corresponding single-modality frameworks (S-S and C-C). This is because the current framework does not consider mutual promotion between different modalities. We will explore this direction in future research.

Table \ref{t1}  shows the optimal results of the five frameworks in Figure \ref{exp} under four different data conditions. For the C-SC, S-SC and SC-SC frameworks, we chose the results corresponding to the optimal mean of the CNN and SNN.

\subsection{Competitive Suppression in SC-SC}
This set of experiments investigates the SC-SC framework by varying the number of clients (denoted as N) and the number of uploads (denoted as P). This variation aims to further explore the issue of competition suppression within the framework. Each configuration of N/P considers both IID and non-IID (with alpha=0.5) scenarios.

Observations from Figure \ref{exp1} are as follows. As the values of N/P increase, the convergence discrepancy between SNN and CNN decreases gradually. In the IID scenario, role reversal between superior and inferior parties is more likely to occur. Initially, in the IID column, CNN clients (blue curve) tend to be dominant, but later, it transitions to SNN clients (orange curve) becoming more dominant. However, non-IID significantly reduces the competition suppression problem in SC-SC federated training.

\section{Conclusion}

In this paper, we investigate a variety of approaches to multi-device federated learning that integrates CNNs and SNNs. To provide a comprehensive understanding of this framework, we conduct a thorough comparison with various existing approaches, including federated CNN, federated SNN, federated CNN for SNN, federated SNN for CNN, and federated CNN with SNN fusion. The experimental results reveal superiority of the SNN-CNN fusion compared with other heterogeneous frameworks and also explore the phenomenon of competition suppression in the SNN-CNN fusion training process.

However, due to the differences and misalignment between modalities, the SNN-CNN fusion framework is slightly worse than than the single-modality framework. In the future, we aim to explore alignment and transfer learning techniques within this fusion framework to further enhance its accuracy and robustness. This future work holds promise for advancing the capabilities of multi-device federated learning and extending its applicability to diverse real-world scenarios.

\bibliographystyle{named}
\bibliography{ijcai24}

\begin{thebibliography}{}

\bibitem[\protect\citeauthoryear{Custers \bgroup \em et al.\egroup }{2019}]{custers2019eu}
Bart Custers, Alan~M Sears, Francien Dechesne, Ilina Georgieva, Tommaso Tani, and Simone Van~der Hof.
\newblock {\em EU personal data protection in policy and practice}.
\newblock Springer, 2019.

\bibitem[\protect\citeauthoryear{Deng}{2012}]{deng2012mnist}
Li~Deng.
\newblock The mnist database of handwritten digit images for machine learning research.
\newblock {\em IEEE Signal Processing Magazine}, 29(6):141--142, 2012.

\bibitem[\protect\citeauthoryear{Ghosh-Dastidar and Adeli}{2009}]{ghosh2009spiking}
Samanwoy Ghosh-Dastidar and Hojjat Adeli.
\newblock Spiking neural networks.
\newblock {\em International journal of neural systems}, 19(04):295--308, 2009.

\bibitem[\protect\citeauthoryear{Gu \bgroup \em et al.\egroup }{2018}]{gu2018recent}
Jiuxiang Gu, Zhenhua Wang, Jason Kuen, Lianyang Ma, Amir Shahroudy, Bing Shuai, Ting Liu, Xingxing Wang, Gang Wang, Jianfei Cai, et~al.
\newblock Recent advances in convolutional neural networks.
\newblock {\em Pattern recognition}, 77:354--377, 2018.

\bibitem[\protect\citeauthoryear{Heeger and others}{2000}]{heeger2000poisson}
David Heeger et~al.
\newblock Poisson model of spike generation.
\newblock {\em Handout, University of Standford}, 5(1-13):76, 2000.

\bibitem[\protect\citeauthoryear{Hunsberger and Eliasmith}{2015}]{hunsberger2015spiking}
Eric Hunsberger and Chris Eliasmith.
\newblock Spiking deep networks with lif neurons.
\newblock {\em arXiv preprint arXiv:1510.08829}, 2015.

\bibitem[\protect\citeauthoryear{Ji \bgroup \em et al.\egroup }{2024}]{ji2024emerging}
Shaoxiong Ji, Yue Tan, Teemu Saravirta, Zhiqin Yang, Yixin Liu, Lauri Vasankari, Shirui Pan, Guodong Long, and Anwar Walid.
\newblock Emerging trends in federated learning: From model fusion to federated x learning.
\newblock {\em International Journal of Machine Learning and Cybernetics}, pages 1--22, 2024.

\bibitem[\protect\citeauthoryear{Johnson and Zhang}{2013}]{johnson2013accelerating}
Rie Johnson and Tong Zhang.
\newblock Accelerating stochastic gradient descent using predictive variance reduction.
\newblock {\em Advances in neural information processing systems}, 26, 2013.

\bibitem[\protect\citeauthoryear{Koo \bgroup \em et al.\egroup }{2020}]{koo2020sbsnn}
Minsuk Koo, Gopalakrishnan Srinivasan, Yong Shim, and Kaushik Roy.
\newblock Sbsnn: Stochastic-bits enabled binary spiking neural network with on-chip learning for energy efficient neuromorphic computing at the edge.
\newblock {\em IEEE Transactions on Circuits and Systems I: Regular Papers}, 67(8):2546--2555, 2020.

\bibitem[\protect\citeauthoryear{Kornijcuk \bgroup \em et al.\egroup }{2016}]{kornijcuk2016leaky}
Vladimir Kornijcuk, Hyungkwang Lim, Jun~Yeong Seok, Guhyun Kim, Seong~Keun Kim, Inho Kim, Byung~Joon Choi, and Doo~Seok Jeong.
\newblock Leaky integrate-and-fire neuron circuit based on floating-gate integrator.
\newblock {\em Frontiers in neuroscience}, 10:212, 2016.

\bibitem[\protect\citeauthoryear{Lobo \bgroup \em et al.\egroup }{2020}]{lobo2020spiking}
Jesus~L Lobo, Javier Del~Ser, Albert Bifet, and Nikola Kasabov.
\newblock Spiking neural networks and online learning: An overview and perspectives.
\newblock {\em Neural Networks}, 121:88--100, 2020.

\bibitem[\protect\citeauthoryear{Lu and Fan}{2020}]{lu2020efficient}
Yanyang Lu and Lei Fan.
\newblock An efficient and robust aggregation algorithm for learning federated cnn.
\newblock In {\em Proceedings of the 2020 3rd International Conference on Signal Processing and Machine Learning}, pages 1--7, 2020.

\bibitem[\protect\citeauthoryear{McMahan \bgroup \em et al.\egroup }{2017}]{mcmahan2017communication}
Brendan McMahan, Eider Moore, Daniel Ramage, Seth Hampson, and Blaise~Aguera y~Arcas.
\newblock Communication-efficient learning of deep networks from decentralized data.
\newblock In {\em Artificial intelligence and statistics}, pages 1273--1282. PMLR, 2017.

\bibitem[\protect\citeauthoryear{Merenda \bgroup \em et al.\egroup }{2020}]{merenda2020edge}
Massimo Merenda, Carlo Porcaro, and Demetrio Iero.
\newblock Edge machine learning for ai-enabled iot devices: A review.
\newblock {\em Sensors}, 20(9):2533, 2020.

\bibitem[\protect\citeauthoryear{Rashid \bgroup \em et al.\egroup }{2022}]{rashid2022ahar}
Nafiul Rashid, Berken~Utku Demirel, and Mohammad~Abdullah Al~Faruque.
\newblock Ahar: Adaptive cnn for energy-efficient human activity recognition in low-power edge devices.
\newblock {\em IEEE Internet of Things Journal}, 9(15):13041--13051, 2022.

\bibitem[\protect\citeauthoryear{Skatchkovsky \bgroup \em et al.\egroup }{2020}]{skatchkovsky2020federated}
Nicolas Skatchkovsky, Hyeryung Jang, and Osvaldo Simeone.
\newblock Federated neuromorphic learning of spiking neural networks for low-power edge intelligence.
\newblock In {\em ICASSP 2020-2020 IEEE International Conference on Acoustics, Speech and Signal Processing (ICASSP)}, pages 8524--8528. IEEE, 2020.

\bibitem[\protect\citeauthoryear{Venkatesha \bgroup \em et al.\egroup }{2021}]{venkatesha2021federated}
Yeshwanth Venkatesha, Youngeun Kim, Leandros Tassiulas, and Priyadarshini Panda.
\newblock Federated learning with spiking neural networks.
\newblock {\em IEEE Transactions on Signal Processing}, 69:6183--6194, 2021.

\bibitem[\protect\citeauthoryear{Wang \bgroup \em et al.\egroup }{2020}]{wang2020federated}
Hongyi Wang, Mikhail Yurochkin, Yuekai Sun, Dimitris Papailiopoulos, and Yasaman Khazaeni.
\newblock Federated learning with matched averaging.
\newblock {\em arXiv preprint arXiv:2002.06440}, 2020.

\bibitem[\protect\citeauthoryear{Wang \bgroup \em et al.\egroup }{2023}]{wang2023efficient}
Yuan Wang, Shukai Duan, and Feng Chen.
\newblock Efficient asynchronous federated neuromorphic learning of spiking neural networks.
\newblock {\em Neurocomputing}, 557:126686, 2023.

\bibitem[\protect\citeauthoryear{Yang \bgroup \em et al.\egroup }{2019}]{yang2019federated}
Qiang Yang, Yang Liu, Tianjian Chen, and Yongxin Tong.
\newblock Federated machine learning: Concept and applications.
\newblock {\em ACM Transactions on Intelligent Systems and Technology (TIST)}, 10(2):1--19, 2019.

\bibitem[\protect\citeauthoryear{Yang \bgroup \em et al.\egroup }{2022}]{yang2022lead}
Helin Yang, Kwok-Yan Lam, Liang Xiao, Zehui Xiong, Hao Hu, Dusit Niyato, and H~Vincent~Poor.
\newblock Lead federated neuromorphic learning for wireless edge artificial intelligence.
\newblock {\em Nature communications}, 13(1):4269, 2022.

\bibitem[\protect\citeauthoryear{Zhang \bgroup \em et al.\egroup }{2023}]{zhang2023federated}
Xunzheng Zhang, Alex Mavromatis, Antonis Vafeas, Reza Nejabati, and Dimitra Simeonidou.
\newblock Federated feature selection for horizontal federated learning in iot networks.
\newblock {\em IEEE Internet of Things Journal}, 10(11):10095--10112, 2023.

\bibitem[\protect\citeauthoryear{Zhu \bgroup \em et al.\egroup }{2021}]{zhu2021federated}
Hangyu Zhu, Jinjin Xu, Shiqing Liu, and Yaochu Jin.
\newblock Federated learning on non-iid data: A survey.
\newblock {\em Neurocomputing}, 465:371--390, 2021.

\end{thebibliography}

\end{document}